\documentclass[letterpaper]{article} 
\usepackage{aaai23}  
\usepackage{times}  
\usepackage{helvet}  
\usepackage{courier}  
\usepackage[hyphens]{url}  
\usepackage{graphicx} 
\def\UrlFont{\rm}  
\usepackage{natbib}  
\usepackage{caption} 
\usepackage{subcaption}
\usepackage{float}
\DeclareCaptionStyle{ruled}{labelfont=normalfont,labelsep=colon,strut=off} 
\usepackage{algorithm}
\usepackage{algorithmic}
\usepackage{adjustbox}

\usepackage{newfloat}
\usepackage{listings}
\DeclareCaptionStyle{ruled}{labelfont=normalfont,labelsep=colon,strut=off} 
\floatstyle{ruled}
\newfloat{listing}{tb}{lst}{}
\floatname{listing}{Listing}
\title{Novel Intent Detection and Active Learning Based Classification (Student Abstract)}
\author{
    Ankan Mullick
}
\affiliations{
    Department of Computer Science and Engineering.
    Indian Institute of Technology Kharagpur, India \\
    ankanm@kgpian.iitkgp.ac.in

}

\usepackage{bibentry}

\begin{document}

\maketitle

\begin{abstract}
Novel intent class detection is an important problem in real world scenario for conversational agents for continuous interaction. Several research works have been done to detect novel intents in a mono-lingual (primarily English) texts and images. But, current systems lack an end-to-end universal framework to detect novel intents across various different languages with less human annotation effort for mis-classified and system rejected samples. This paper proposes \textbf{NIDAL} (\textbf{N}ovel \textbf{I}ntent \textbf{D}etection and \textbf{A}ctive \textbf{L}earning based classification), a semi-supervised framework to detect novel intents while reducing human annotation cost. Empirical results on various benchmark datasets demonstrate that this system outperforms the baseline methods by more than 10\%  margin for accuracy and macro-F1. The system achieves this while maintaining overall annotation cost to be just $\sim$ 6-10\% of the unlabeled data available to the system.
\end{abstract}


\section{Introduction} 
\label{sec:introduction}
With the emerging new intents in the system, the conversational agents need to be retrained in order to identify these newer intents aka \textit{novel classes}. Also as time progresses, the overall accuracy of identifying the known intents for the user queries should not be compromised. Currently, if a user utterance belongs to one of the known intents but the confidence score is low, these are called \textit{rejected utterances}. It is infeasible to label all the new instances because of the sheer volume, and hence an effective mechanism to reduce human annotation cost would be required. To address these issues, in this work, a semi-supervised setting is adopted to identify known and novel intent classes. Corresponding to this problem setting, it starts with an initial labelled training data which consists of a set of known intents and a large unlabelled data that comprises of known and novel intent classes. The evaluation is performed on a predefined test set that consists of all the known and novel classes.

In the last decade, researchers have focused on out of domain class detection and explored various active learning methods in various scenarios to improve classification and lowering the human annotation effort.
 \textbf{Novel Class Detection:}  
Some advanced approaches are developed to explore class emergence problem in data stream like SEEN~\cite{zhu2020semi}, Zero-Shot-OOD~\cite{tan2019out} and SENC-MaS~\cite{mu2017streaming} but there is less focus on novel intent detection. \textbf{Active Learning:} ModAL \cite{danka2018modal} propose various active learning (AL) strategies to improve system accuracy. \cite{tian2011active} develop a maximal marginal uncertainty based AL model. But, these are not focused on rejected utterances. 

In this work, an end-to-end framework is proposed to identify novel intents with increased in-domain class detection accuracy while handling system rejected utterances.

\section{Dataset} 
\label{sec:dataset}

NIDAL approach is compared with similar models and evaluated across several standard public dataset in NLU domain - SNIPS~\cite{coucke2018snips}, ATIS~\cite{tur2010left} and Facebook Multi-lingual \cite{schuster2018cross}. 
This framework is language agnostic and performs significantly well on all the datasets.

\begin{table*}[!t]
\centering
\begin{adjustbox}{width=0.85\linewidth}
\begin{tabular}{|c|c|c|c|c|c|c|c|c|c|c|}
\hline
\textbf{Method} & \multicolumn{2}{c|}{\textbf{FB-EN}} & \multicolumn{2}{c|}{\textbf{FB-ES}} & \multicolumn{2}{c|}{\textbf{FB-TH}} & \multicolumn{2}{c|}{\textbf{ATIS}} &\multicolumn{2}{c|}{\textbf{SNIPS}} \\\cline{2-11} 
                                 & \textbf{Acc}    & \textbf{Mac F1}   & \textbf{Acc}    & \textbf{Mac F1}   & \textbf{Acc}    & \textbf{Mac F1}   & \textbf{Acc}    & \textbf{Mac F1}
                                   & \textbf{Acc}    & \textbf{Mac F1} 
                                 \\ \hline
SEEN*                           & 84.9            & 86.2              & 75.4            & 79.8              & 71.9            & 77.5              & 84.3            & 71.4 & 85.8 & 86.2            \\ \hline
Zero-Shot-OOD*                        & 86.7            & 87.4              & 62.1            & 70.1              & 68.6            & 72.1              & 84.8            & 81.7 & 87.4 & 88.5             \\ \hline
SENC-MaS* & 82.3 & 83.9 & 69.4 & 70.5 & 64.6 & 72.8 & 80.1 & 69.4 & 82.0 & 81.4\\\hline

NIDAL &  \textbf{98.0} & \textbf{93.1} & \textbf{91.5} & \textbf{96.1} & \textbf{92.5} & \textbf{89.5} & \textbf{92.1} & \textbf{88.9} & \textbf{97.9}& \textbf{95.5} \\\hline

\end{tabular}
\end{adjustbox}
\caption{Overall Accuracy (Acc) and Macro average F1-score on Facebook-English, Spanish, Thai, ATIS and SNIPS Datasets. * - advantage to these baselines, since they do not distinguish between multiple novel intents.}
\label{tab:fb_results}
\vspace{-5mm}
\end{table*}

\section{Approach}

The framework ``NIDAL'' is divided into two major components - novel intent detection and active learning. The Novel Intent Detection module detects out-of-domain samples on unlabelled data and the active learning based known intent classification, handles the rejected utterances ensuring higher overall accuracy on the known intents. 

\noindent \textbf{Neural Model} ($\mathcal{M}$): The JointBERT \cite{chen2019bert} is used as the learning model for intent classification.
English uncased BERT-Base model is used 
as the base model for JointBERT. For other languages, bert-base-multilingual-uncased model is used. The best results can be obtained when the model is trained for 10 epochs and the learning rate is $5e-5$.


\noindent \textbf{Part 1: Novel Intent Detection (NID):} \label{sec:ood}
 The algorithm for Novel Intent Detection consists of the following Steps:
\textbf{Step 1 - \textit{Cycle 0}}: With the initial labelled data ($\mathcal{L}$), consisting of examples only from the known intent set, a Neural Model ($\mathcal{M}$) is trained. 
    \textbf{Step 2 - \textit{Out-Of-Domain Sample Detection (OOD-SD)}}: MSP \cite{hendrycks2018baseline} 
algorithm is considered to detect Out-Of-Domain samples $\mathcal{U'}$ on the Unlabelled Data ($\mathcal{U}$). \textbf{Step 3 - \textit{Labeling of Novel Intents}}: $m\%$ of the OOD samples is considered $\mathcal{U'}$ for manual annotation ($\mathcal{U}'_{m}$). Experiments are done with various labeling strategies to add the rest of the OOD samples back to Cycle 0 training set ($\mathcal{L}$). 
    \textbf{Step 4 - \textit{Cycle 1}}: $\mathcal{M}$ is re-trained on the modified training set to perform prediction on the Test Set. 


\noindent \textbf{Part 2: Active Learning (AL):} \label{sec:alc}
Next, an active learning based framework is devised to take care of rejected utterances with a low confidence. 
The algorithm is as follows:
\textbf{Step 1 - \textit{Cycle 1}}: The retrained Neural Model ($\mathcal{M}$) is used to predict intent scores on the unlabelled dataset after removing OODs ($\mathcal{U}_{rem}$).  
\textbf{Step 2 - \textit{Cycle 1}}: A pre-defined threshold ($\mathcal{TH}$) is set, and the samples with score  $< \mathcal{TH}$ are then fed XGBoost (XGB) classifier and the prediction score is used to identify their labels (\textit{auto-corrected}). Samples which are rejected again, are manually annotated. These auto-corrected and annotated samples are added back to the labelled data and removed from the unlabelled data. \textbf{Step 3 - \textit{Cycle 2}}: The Neural Model ($\mathcal{M}$) is re-trained with updated labelled dataset and redo step 1-2 for the same threshold value. \noindent \textbf{Step 4:} This approach is run for $2\leq K\leq 5$ cycles.   

\section{Experimental Results}

Softmax Prediction Probability (MSP)
is used to predict out-of-domain samples based on the softmax prediction scores. Different threshold values on top of Neural Model ($\mathcal{M}$) is set to identify rejected utterances and optimum results are obtained when threshold is set to 75\% of maximum classification probability score for a particular dataset. These rejected utterances are passed through XGBoost (XGB) classifier for auto-correction. The rest of the samples (i.e. XGB rejected) are annotated manually. These auto corrected and manually annotated samples are removed from the unlabelled set and added back to labelled dataset, $L$ for retraining purposes. Thus the auto-correction criteria, saves significant human labeling cost.

The results for Novel Intent Detection and Active Learning (NIDAL) in terms of accuracy and macro f1 score for various english (Facebook-English [FB-EN], SNIPS, ATIS) and non-english data (Facebook-Thai [FB-TH] and Facebook-Spanish [FB-ES]) are shown in Table \ref{tab:fb_results}.

\noindent \textbf{Competing Baselines:}
Experiments are done with modified versions of various existing methods to compare NIDAL - SEEN~\cite{zhu2020semi}, Zero-Shot-OOD~\cite{tan2019out} and SENC-MaS~\cite{mu2017streaming}. Experiments are done with different variations of baselines (parameter and hyperparameter tuning) and the best results are reported.
Since, all these methods can only detect one novel class at a time, when considering the multiple novel intents in test set as a single novel class for these methods and report the accuracy and macro-F1 considering the various known classes and a novel class for the same. 
The results for different baselines are shown in Table \ref{tab:fb_results}. It is observed that the annotation cost varies from 6-10\% samples of the unlabelled data. Considering the huge gains in performance, this is a reasonable annotation cost trade-off. These experiments show that given the same amount of human labeling, this framework works much better than the baselines.
 
\section{Conclusion} 
\label{sec:conclusion}
This paper proposes NIDAL, an end-to-end framework for detection of novel intents, as well as to handle system rejected utterances using an active learning framework. Experiments are performed on various benchmark datasets and the results consistently show the efficacy of the proposed framework. Specifically, it achieves large and consistent gains in performance with a small human annotation cost across different datasets. 
One future step is to extend this work on diverse domains (like finance, technology, healthcare etc.).

\bibliography{main}

\end{document}